\documentclass[submission,copyright,creativecommons]{eptcs}
\providecommand{\event}{SLPCS 2016} % Name of the event you are submitting to

\usepackage{scrextend}
\usepackage[utf8]{inputenc}
\usepackage[english]{babel}
\usepackage{amsmath}
\usepackage{amsfonts}
\usepackage{amsthm}
\usepackage{graphicx}
\usepackage{subcaption}
\usepackage{wrapfig}
\usepackage[toc,page]{appendix}

\newtheorem{theorem}{Theorem}
\newtheorem{definition}{Definition}
\newcommand{\sleq}{\sqsubseteq}

\title{Entailment Relations on Distributions}
\author{John van de Wetering
\institute{Radboud University\\Nijmegen, Netherlands}
\institute{Department of Computer Science\\
Oxford University\\
Oxford, United Kingdom}
\email{j.wetering@student.ru.nl}
}
\def\titlerunning{Entailment Relations on Distributions}
\def\authorrunning{J. van de Wetering}

\begin{document}

\maketitle

\begin{abstract}
In this paper we give an overview of partial orders on the space of probability distributions that carry a notion of information content and serve as a generalisation of the Bayesian order given in \cite{coecke2010book}. We investigate what constraints are necessary in order to get a unique notion of information content.
These partial orders can be used to give an ordering on words in vector space models of natural language meaning relating to the contexts in which words are used, which is useful for a notion of entailment and word disambiguation. The construction used also points towards a way to create orderings on the space of density operators which allow a more fine-grained study of entailment. The partial orders in this paper are directed complete and form domains in the sense of domain theory.
\end{abstract}

\section{Introduction}
Distributional models of natural language form a popular way to study language in the context of automated natural language processing. These models rely on the Distributional Hypothesis: words that occur in similar contexts have similar meanings. 

The categorical compositional distributional model of natural language meaning developed by Coecke, Sadrzadeh and Clark \cite{coecke2010,clark2008} also gives a way of composing distributions. It has been corroborated empirically that on some tests this model performs better then the state of the art \cite{kartsaklis2012}.

Recently this model has been expanded to take into account lexical ambiguity, notions of homonymy and polysemy \cite{piedeleu2015} and entailment at the word and sentence level by passing from a vector space model to a density matrix model \cite{balkir2015,balkir2016,bankova2016}. In these papers various relations between the density matrices were explored to get a definition of entailment on the distributional level, such as the fidelity and the relative entropy. In \cite{bankova2016} a modified Löwner ordering was used to get a notion of graded entailment. In \cite{kotlerman2010} they constructed a nonsymmetric similarity measure based on a modified measure of feature inclusion.

In any of these cases the goal is to get a relation between pairs of words or sentences that captures the idea of information content. We say that the word \emph{dog} entails the word \emph{animal} because in most contexts where the word \emph{dog} is used, we could use the more general (less specific, less informative) word \emph{animal}.

The same is true for word disambiguation. Consider the word \emph{bank}. It might mean \emph{river bank} or \emph{investment bank}. Without any further context we don't know which one is meant. The word \emph{bank} offers less information than either of these more specific words. We can consider it to be in a mixed state of these pure meanings, which collapses to a pure state when given the right context.

It therefore makes sense that to get a notion of disambiguation or lexical entailment we should be looking for a relation that captures the idea of information content. The obvious properties that we would require of such a relation are those of a partial order: reflexive, transitive and antisymmetric.

The way word vectors are usually constructed is by counting the coocurrence with some set of basis words. The components can then be interpreted as probabilities of a word occuring at the same time as this specific basis word. So in fact, the word vector can be seen as a probability distribution.

If a word is instead represented by a density matrix then when it is diagonalised we have a probability distribution on the diagonal. This means that a relation capturing the notion of information content should at least be a partial order on the space of probability distributions.

There is a well known partial order on the space of positive semi-definite matrices called the Löwner order, but on the space of density operators no two different density operators are comparable (if we have $x\sleq y$ then we must have $x=y$). This is a direct effect of the normalisation of the trace of the operators. A modification to the Löwner order was made in \cite{bankova2016} in order to get a notion of graded entailment. The resulting structure was no longer a partial order, since the modification removed transitivity and replaced it with a weaker condition. In this paper we will show two different modifications to the Löwner order that do result in proper partial orders.

An example of a nontrivial partial order on the space of probability distributions that has suitable information-like properties is the Bayesian order outlined in \cite{coecke2003,coecke2010book}. This is in fact the only example the author could find in literature. The Bayesian order served as the inspiration for this paper and the results outlined here can be seen as generalisations of the results related to the Bayesian order.

In this paper we will explore what conditions we need in order for the resulting partial order to represent information content. We will also look at what kind of conditions we need in addition to get a unique notion of information content. Since there has been surprisingly little work in the area of partial orders representing information we will focus on partial orders on probability distributions instead of on the bigger space of density operators. We will also just be looking at entailment on the word level and leave compositionality for further research.

Note also that the results in this paper might prove useful in resource theory and quantum information theory as density matrices are quantum states and probability distributions are classical states. The partial orders studied in this paper turn out to be domains: directed complete partial orders which are exact.

\section{Background}
We begin by stating the definition of a partial order.
\begin{definition}
\em
A \emph{partial order} on a space $S$ is a binary relation $\sleq$ which is
\begin{enumerate}
\item Reflexive: $\forall x \in S: x \sleq x$.
\item Transitive: $\forall x,y,z \in S: x \sleq y \text{ and }y \sleq z \implies x\sleq z$.
\item Antisymmetric: $\forall x,y \in S: x\sleq y \text{ and } y\sleq x \implies x=y$.
\end{enumerate}
\end{definition}
We can restrict a partial order on the density matrices to the diagonal density matrices. This is equivalent to the space of finite probability distributions $\Delta^n = \{(x_1,\ldots,x_n) ; x_i \geq 0, \sum_i x_i = 1\}$, which can be interpreted geometrically as the $(n-1)$-simplex.

We can then wonder when this procedure can be reversed: which partial orders on the diagonal density matrices extend to a partial order on the entire space of density matrices? The naive approach is to define $\rho \sleq^\prime \pi$ iff Diag$(\rho)\sleq$Diag$(\pi)$, where Diag$(\rho)$ is the probability distribution of the eigenvalues of $\rho$. However if we take an arbitrary density matrix $\rho$ the diagonalisation will not be fully determined: we can still freely permute the basis vectors. Reflexivity would then imply that any permutation of basis must be equivalent which would in turn break antisymmetry. We must require that $\rho$ and $\pi$ be diagonalised simultaneously in order for them to be comparable by a partial order on $\Delta^n$.

If two density matrices can be diagonalised simultaneously then there is still a freedom of permuting the basis vectors, so a necessary condition for $\sleq$ to be extended to the entire space of density operators is for it to be invariant under basis vector permutation:
\begin{definition} \em
Let $\sleq$ be a partial order on $\Delta^n$. We call it \emph{permutation invariant} if for any permutation $\sigma\in S^n$: $x\sleq y \implies \sigma(x)\sleq \sigma(y)$
\end{definition}
It can be shown that a permutation invariant partial order extends to a partial order on the density matrices (a density matrix is completely determined by its eigenvalues and an orthonormal basis). 

A notion of information content is Shannon entropy. On $\Delta^n$ the element with the highest amount of entropy is the uniform distribution $\bot = \frac{1}{n}(1,\ldots,1)$. The elements with the lowest amount of entropy are the pointed distributions that have $x_i=1$ for some $i$ and the rest equal to zero, also called the `pure' states. Denote these as $\top_i$. Intuitively $\bot$ is the element with the lowest amount of information, and $\top_i$ are the elements with the most amount of information. We require that our partial order on $\Delta^n$ respects this: every distribution contains more information than $\bot$ and every distribution is smaller than at least one maximal element.\\
Linguistically a word would be represented by the uniform distribution if it occured the same amount of times in any context, but such a word would of course not add any information to the sentence. A candidate for such a word would for instance be `\emph{the}'. Realistically no word will be represented by the uniform distribution, but we would find examples of words that are uniformly distributed on a subset of contexts. Such as the word `\emph{bank}' that we would expect somewhat uniformly in the contexts of finance and rivers. Stating that each word can be compared to some pure state is akin to stating that each word can be resolved to some pure meaning.

In order to restrict ourselves to nontrivial partial orders we will require one further property: that the partial order respects the mixing of information content, defined as such:
\begin{definition}\em
We say that a partial order on $\Delta^n$ allows \emph{mixing} when we have for any $x,y$ and $t\in[0,1]$:
\begin{equation*}
    x\leq y \implies x\leq (1-t)x+ty \leq y
\end{equation*}
\end{definition}
This states that when an element contains less information than another and this information is comparable, then mixing the information content will give something with an information content in between. Note that the space of probability distributions is convex. This demand makes the partial order respect that convexity in a natural way. We are now ready to give a minimal definition of partial order that represents information content.

\begin{definition} \em
A partial order on $\Delta^n$ which is permutation invariant, allows mixing and has the uniform distribution as the minimal element and the pointed distributions as the maximal elements is an \emph{information ordering}.
\end{definition}

\begin{wrapfigure}{r}{4.5cm}
\includegraphics[trim={0 0.3cm 0 0.5cm}]{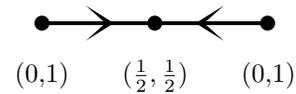}
\caption{The unique partial order on $\Delta^2$ satisfying Definition 4.}
\vspace{-0.5cm}
\end{wrapfigure}

There is a unique partial order satisfying the conditions of Definition 4 on $\Delta^2$ as seen in Figure 1. The pure distributions are at the ends while the uniform distribution is in the middle.

We might hope that these conditions also uniquely determine a partial order for higher values of $n$, but this is not the case. The inductive procedure in \cite{coecke2010book} uniquely determines a partial order that does have the right properties, but as we will see we can create other partial orders without using this inductive procedure. The structure that \emph{is} fixed by these conditions is illustrated for $n=3$ in Figure 2.

\begin{wrapfigure}{l}{0.35\textwidth}
\includegraphics[trim={0 0.5cm 0 0.5cm}, width=0.35\textwidth]{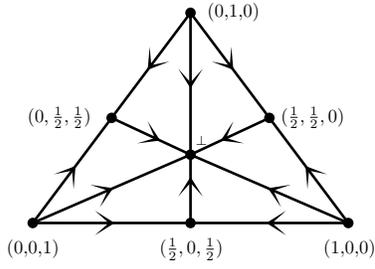}
\caption{The basic structure of an information ordering on $\Delta^3$.}
\vspace{-0.5cm}
\end{wrapfigure}

We see that the space is cut up into natural regions. We will refer to these as sectors, and will come back to those later.

For reference we will state a definition of the Bayesian order here. The other partial orders in this paper will have a similar format.

\begin{definition} \em
The Bayesian order $\sleq_B$ is defined as
$x\sleq_B y$ iff there is a permutation $\sigma$ such that the coordinates of $\sigma(x)$ and $\sigma(y)$ are both monotonically decreasing and we have\\
$ \sigma(x)_i \sigma(y)_{i+1} \leq \sigma(y)_i \sigma(x)_{i+1}~$ for all $1\leq i \leq n-1$. \cite{coecke2010book}
\end{definition}
The condition that comparable elements must both be able to be permuted in the same way might be seen as odd, but it in fact ensures that the elements are part of the same sector (one of the smaller triangles in Figure 2). As we will see in Section 4, the Bayesian order belongs to a class of partial orders that have this property.

\section{Non-Uniqueness of Information Orderings}
We will start by showing that the requirements of Definition 4 are not strong enough to give a unique definition of information content. That is: there exist partial orders $\sleq_1$ and $\sleq_2$ such that there are points $x\neq y$ with $x\sleq_1 y$ but $y\sleq_2 x$. 

\subsection{Renormalising the Löwner order}
As stated in the introduction, the Löwner order given by $x\sleq y$ iff $y-x\geq 0$ is trivial ($x\sleq y \implies x=y$) on $\Delta^n$. This is due to the fact that the components of $x$ and $y$ both need to sum up to 1. By renormalising the components so that they no longer sum up to the same value, we are able to create a nontrivial order.

There are at least two natural choices for renormalisation: we can set the largest coordinate equal to 1, or we can set the smallest coordinate equal to 1. 

The normalisation to the largest coordinate gives the partial order
\begin{equation*}
    x\sleq_L^+ y \iff x^+y_k \leq y^+x_k \text{ for all }k.
\end{equation*}
where $x^+$ is defined as $x^+=$max$\{x_k\}$. This partial order satisfies all the conditions specified in Definition 4, so it is an information ordering.

The normalisation to the smallest coordinate is slightly more difficult since the smallest coordinate could be equal to zero. If both elements have the same amount of zeroes we can ignore those and use the smallest nonzero element. If an element $y$ has strictly more zeroes than $x$ we can view $y$ as being blown up to infinity while $x$ stays finite, so we would simply define $x\sleq y$, as long as their common zeroes are in the same positions. 

\noindent Keeping this in mind, we can define the second renormalised Löwner order by induction on $n$ as $x \sleq_{L,n}^- y$ if and only if one of the following holds:
\begin{enumerate}
\item There is a $k$ such that $x_k = 0$, $y_k = 0$ and $x \sleq_{L,(n-1)}^- y$.
\item There is a $k$ such that $y_k = 0$, $x_k = x^-$.
\item For all $k$:  $y_k,x_k\neq 0$ and $x_ky^- \leq y_kx^-$.
\end{enumerate}
Here $x^-$ is defined as the smallest nonzero coordinate.

This is a well-defined partial order and it satisfies all the conditions specified in Definition 4.
In Figure 3 you can see that these two renormalisations make a big difference to the resulting partial order. If we take the points $x=\frac{1}{10}(6,2,2)$ and $y=\frac{1}{30}(15,10,5)$, then $x\sleq_L^+ y$ and $y\sleq_L^-x$. So there is at least one pair of points where $\sleq_L^+$ and $\sleq_L^-$ contradict each other. The conditions specified in Definition 4 are not strong enough to get a unique notion of information content.

\begin{figure}[htb!]
    \centering
    \begin{subfigure}[b]{0.2\textwidth}
    \includegraphics[width=\textwidth]{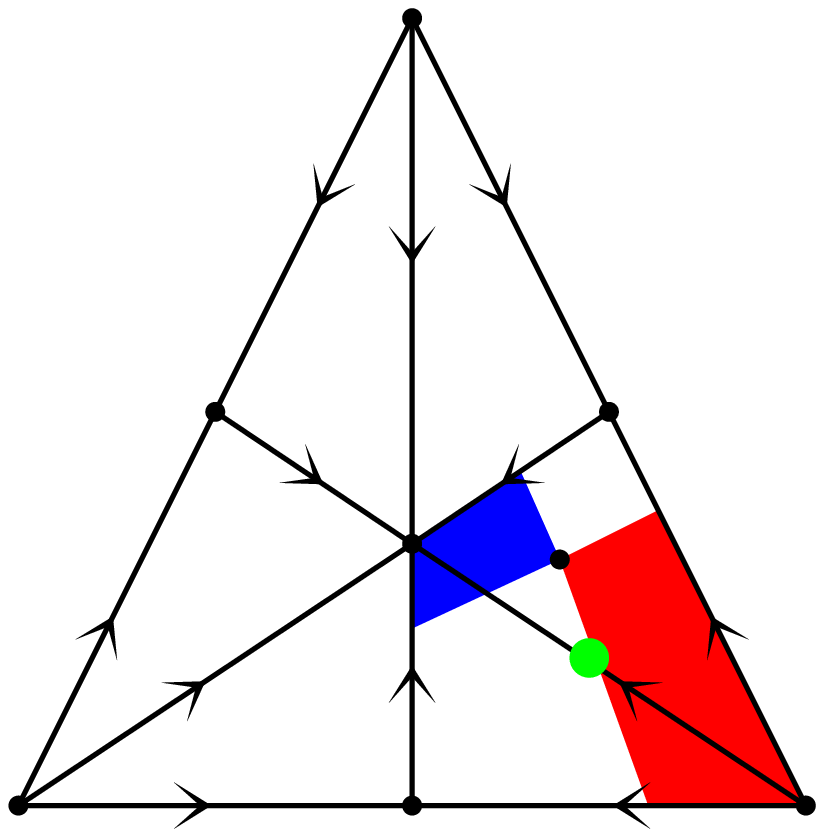}
    \caption{$\sleq_L^+$}
    \end{subfigure}
    ~
    \begin{subfigure}[b]{0.2\textwidth}
    \includegraphics[width=\textwidth]{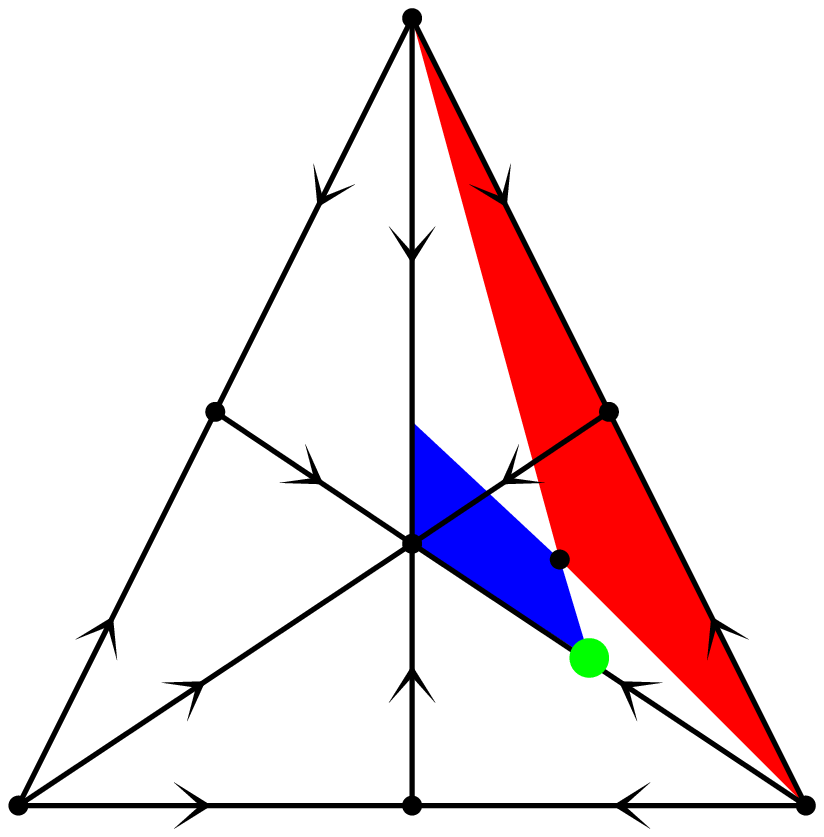}
    \caption{$\sleq_L^-$}
    \end{subfigure}
    \caption{Upperset (red) and downset (blue) of the distribution $y=\frac{1}{30}(15,10,5)$ with respect to the renormalised Löwner orders. The point $x=\frac{1}{10}(6,2,2)$ is denoted in green.}
\end{figure}

\subsection{Measurements prevent contradictions}
A very useful tool to study the relation between different partial orders are \emph{measurements}, the definition of which we take from \cite{coecke2010book} and \cite{martinphd}.
\begin{definition}
A \emph{measurement} is a Scott-continuous strict monotonic map $\mu: (P,\sleq_P) \rightarrow (S,\sleq_S)$.
\end{definition}
Monotonicity means that when $x\sleq_P y$ we have $\mu(x)\sleq_S\mu(y)$ and strictness states that when $x\sleq_P y$ and $\mu(x)=\mu(y)$ we have $x=y$. Scott-continuity is not important for us, but it is a useful property in relation to proving that a partial order is directed complete. All the strict monotonic maps in this paper are also Scott-continuous.

Define the monotone sector of $\Delta^n$ as $\Lambda^n = \{x;x_k\geq x_{k+1}\}$. This corresponds to the lower rightmost triangle in Figure 2. For each $x\in\Delta^n$ there is a unique $y\in\Lambda^n$ such that $y=\sigma(x)$ for some permutation $\sigma$. This gives us a natural retraction $r:(\Delta^n,\sleq)\rightarrow(\Lambda^n,\sleq_{\lvert \Lambda^n})$. $r$ is a measurement for any information ordering $\sleq$ on $\Delta^n$. This means that if we have a measurement $\mu:\Lambda^n \rightarrow S$ this extends to a measurement of $\Delta^n$ by composition with $r$. 

The measurements we will be using are of the form $\mu: \Lambda^n \rightarrow [0,\infty)^*$.\footnote{Any partial order that allows such a measurement is a dcpo \cite{martinphd}.} $[0,\infty)^*$ is the positive interval with the reversed order, so monotonicity means that $x\sleq y$ implies $\mu(x)\geq \mu(y)$. 

The order $\sleq_L^+$ has the measurement $\mu_+(x) = 1 - x^+$. The order $\sleq_L^-$ has a slightly more complicated measure. Define the zero counting function $Z(x)=\#\{k; x_k=0\}$. Then when $x\sleq_L^- y$ we have $Z(y)\geq Z(x)$. If $x\sleq_L^- y$ and $Z(x)=Z(y)$, then $x^-\geq y^-$, and if additionally $x^- = y^-$ then $x=y$. Putting this together we see that $\mu_-(x) = 2n-3 - 2Z(x)) + x^-$ is a measurement of $\sleq_L^-$. We can read this as first counting the amount of zeroes, and then looking at the lowest coordinate. The constant $2n-3$ is added such that $\mu(x) = 0$ iff $x\in$Max$(\Delta^n)$.

These two measurements capture different ideas of what we ``care'' about in our information ordering. Respecting the $\mu_+$ measurement states that the head of a distribution is important, while respecting $\mu_-$ means we care about the tail of a distribution.

Suppose we have two partial orders $\sleq_1$ and $\sleq_2$ that have the same measurement $\mu$. Then if $x\sleq_1 y$ and $y\sleq_2 x$ we get $\mu(x)=\mu(y)$ which gives $x=y$. So partial orders with the same measurement can't contradict each other\footnote{Note that two partial orders that do not have the same measurement don't necessarily have to contradict each other.}. This gives us a tool for ensuring a class of partial orders won't contradict each other.

\section{Restricted Information Orders}
We can extend an information order on $\Delta^n$ to one on the density operators by allowing comparisons if two density operators can be diagonalised simultaneously. Since we have a measurement from $\Delta^n$ to $\Lambda^n$ we can wonder if we can do the same sort of procedure for transitioning from an information order on $\Lambda^n$ to one on $\Delta^n$. That is: we allow comparisons when two elements in $\Delta^n$ can be brought to $\Lambda^n$ simultaneously by some permutation $\sigma(x)$. So in that case $\sigma(x),\sigma(y)\in \Lambda^n$ and we proceed with comparing $\sigma(x)$ and $\sigma(y)$ using a partial order on $\Lambda^n$. This does however not always result in a valid partial order on $\Delta^n$: Suppose we have $x\sleq y$ where $x$ is a border element of $\Lambda^n$. Then it also lies in a neighbouring sector. Suppose there is an element $w$ in this neighboring sector such that $w\sleq x$. Then by transitivity $w\sleq y$. But $w$ and $y$ are in different sectors. So this is a contradiction. A necessary and sufficient condition to prevent this and ensure we can extend a partial order on $\Lambda^n$ to $\Delta^n$ is the following
\begin{definition} \em
A partial order $\sleq$ on $\Lambda^n$ (or $\Delta^n$) satisfies the \emph{degeneracy condition} when for all $x,y\in\Lambda^n$ (or $\Delta^n$) where $x\sleq y$ and $y_i=y_j\neq 0$ we have $x_i=x_j\neq 0$.
\end{definition}
We call this property the degeneracy condition as it ensures that border elements, elements with a degenerated spectrum, are not above any nondegenerated elements. There is a one-to-one correspondence between information orders satisfying the degeneracy condition on $\Lambda^n$ and those orders on $\Delta^n$. We will call an information order that satisfies the degeneracy condition a \emph{restricted} information order as comparisons between elements are restricted to within sectors. The renormalised Löwner orders are not restricted information orders, while the Bayesian order \emph{is} a restricted information order.

We are interested in information-like properties of a distribution $x\in\Delta^n$. If we suppose that all these features can be encoded in terms of real numbers, this would give rise to a feature vector $F(x)$ that is an element of $\mathbb{R}^k$ for some $k$. Comparing the information content of distributions is then translated to comparing the feature vectors of the distributions: $x\sleq y$ iff $F(x)\leq F(y)$ where $\leq$ is the standard product order on $\mathbb{R}^k$: $v\leq w$ iff $v_i\leq w_i$ for all $i$. For instance, for $\sleq_L^+$ the feature vector components are $F(x)_i = x^+/x_i$. For the Bayesian order the feature vector is $F(x)_i = x_i/x_{i+1}$ and for majorization it would be $F(x)_i = \sum_{j=1}^i x_j$. We can classify these types of orders.

\begin{theorem}\em \textbf{Classification of Restricted Information Orderings}. All restricted information orderings of the form $x\sleq y \iff F(x)\leq F(y)$ for some function $F:\Delta^n\rightarrow \mathbb{R}^k$ can be written as the join or meet of the set of partial orders $\sleq_A$ defined as
\begin{align*}
    x\sleq_A y \iff f_i(x)g_i(y)&\leq f_i(y)g_i(x) \text{ for all }1\leq i\leq n-1 \\
    \text{where } f_i(x) &= x_i-x_{i+1} \\
    \text{and } g_i(x) &= y_{i+1} + \sum_{j=i+2}^n A^i_j y_j \text{ where } 1+\sum_{j=i+2}^k A^i_j > 0 \\
    \text{for }&2\leq i \text{ and } i+1<k\leq n \\
    \text{and } g_1(x) &= y_2 + A^1_0 + \sum_{j=3}^n A^1_j y_j \text{ where } 1+2A^1_0 > 0 \\
    \text{and } 1+k&A^1_0+\sum_{j=3}^k A^1_j > 0 \text{ for }2<k\leq n.
\end{align*}
Furthermore, all these partial orders allow $\mu_-$ as a measurement, which means they are all dcpo's. The space of these restricted orders is a complete lattice
\end{theorem}
\smallskip
Note that the feature vectors of these partial orders are $F_i(x) = f_i(x)/g_i(x)$. Using $f_i$ and $g_i$ instead of $F_i$ turns out to be easier because we can deal more naturally with possible zeroes in $g_i$.

We see that all the parameters $A^i_j$ are bounded from below, but not from above. In general, higher values for the parameters correspond to partial orders that are less strict. The Bayesian order is retrieved when setting all parameters to zero. In general, the restricted orders don't respect the ordering given by Shannon entropy. It can be shown that the subset of restricted orders that allow $\mu_+$ as an additional measurement have Shannon entropy as a measurement as well. All the partial orders seen above also have the property that if $x \sleq y$ and $x_k=0$ then $y_k=0$. Or in other words: the \emph{support} of $y$ is included in $x$. This ensures that the relative entropy between $x$ and $y$ is finite. These partial orders are therefore somewhat comparable with the entailment relation of \cite{balkir2015}.

Sharing $\mu_-$ as a measurement ensures that these partial orders don't contradict each other. So the degeneracy condition is a sufficient condition to get a unique direction of information content. Because this space of orderings is a complete lattice there is a unique minimal order and a unique maximal order. The difference between these and the Bayesian order is shown in Figure 4.

\begin{figure}[htb!]
    \centering
    \begin{subfigure}[b]{0.25\textwidth}
    \centering
        \includegraphics[width=0.7\textwidth]{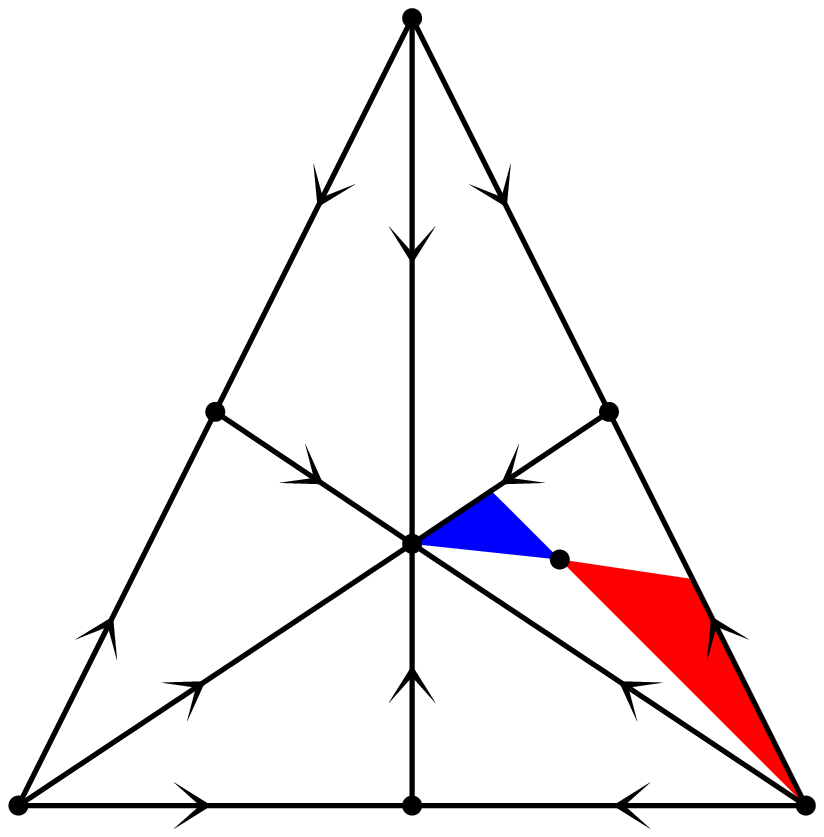}
        \caption{The minimal order}
    \end{subfigure}
    ~
    \begin{subfigure}[b]{0.25\textwidth}
    \centering
        \includegraphics[width=0.7\textwidth]{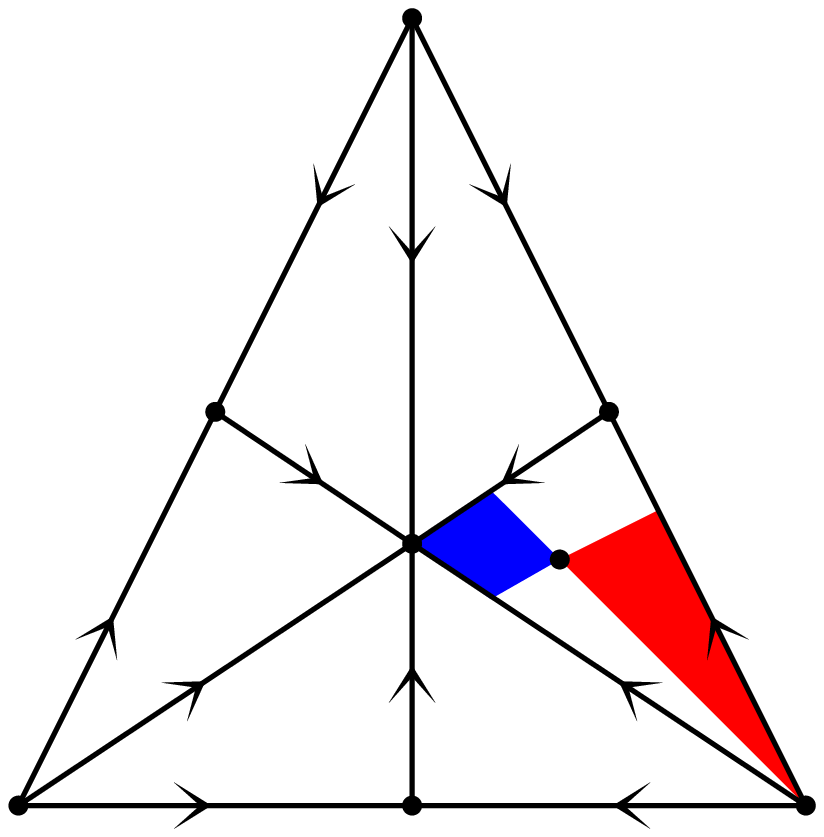}
        \caption{The Bayesian order}
    \end{subfigure}
    ~
    \begin{subfigure}[b]{0.25\textwidth}
    \centering
        \includegraphics[width=0.7\textwidth]{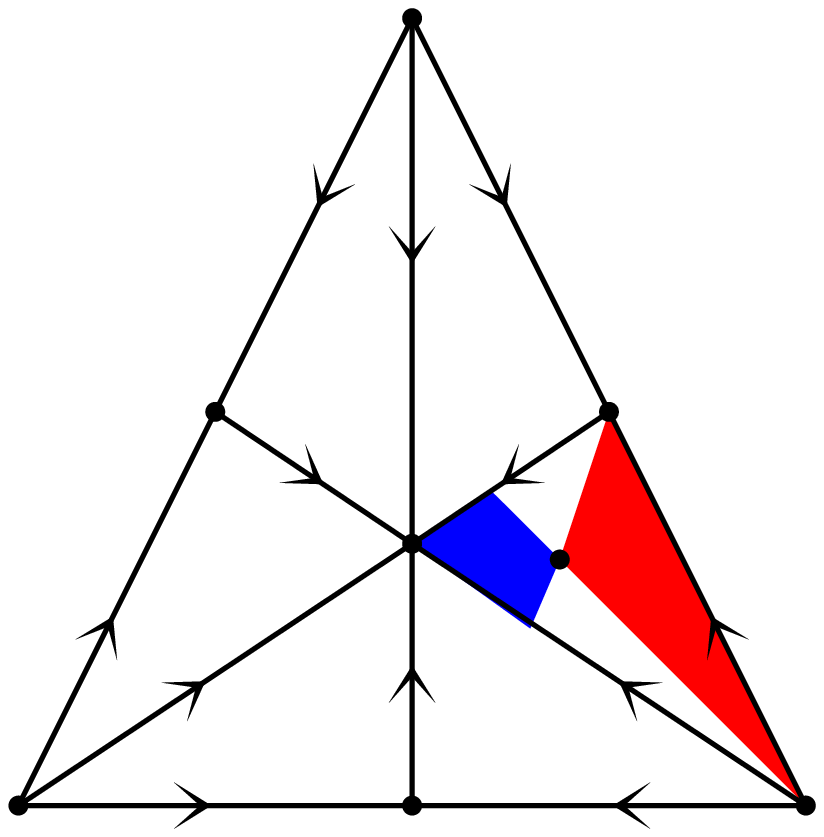}
        \caption{The maximal order}
    \end{subfigure}
    \caption{Upperset (red) and downset (blue) of the same element using the minimal, Bayesian and maximal order on $\Delta^3$.}
\end{figure}

Note that all the restricted orders and $\sleq_L^-$ share the measurement $\mu_-$, so they don't contradict each other. It can also be shown that $\sleq_L^+$ doesn't contradict any restricted order, so both these renormalisations can serve as valid extensions of the restricted orders.

Such an extension is probably necessary. Restricted information orders only allow comparisons within sectors. The amount of sectors in $\Delta^n$ is equal to the factorial of $n$. In an empirical natural language model we would usually have $n$ in the hundreds or thousands, so then there are many more sectors than there are words. This probably means that restricted information orders are too restrictive to be used in practice in the study of natural language as each word will be in its own sector. The renormalised Löwner orders might be better suited to the task.

These partial orders are all constructed by comparing soms feature vectors with each others. This allows for a natural modification to support graded entailment. Suppose we have the partial order $x \sleq y \iff F(x)\leq F(y)$, then we kan define the $p$-entailment as $x \sleq_p y \iff pF(x)\leq F(y)$ for some number $0\leq p \leq 1$. This is no longer a partial order, but a nonsymmetric entailment measure. This generalises the idea of \cite{bankova2016}.

\section{Information Orders on Density Operators}
The central idea behind classifying information orders on probability distributions is that we transition to a feature vector. Let's look at this more closely. We have $\Delta^n\subseteq \mathbb{R}^k$. $\mathbb{R}^k$ has a natural partial order, the product order, which is trivial on $\Delta^n$. By using a feature map $F$ to transform $\Delta^n$ to a different subset of $\mathbb{R}^k$, we can make this partial order nontrivial.

The same sort of procedure can be used on the space of density operators $DO(n)$. This space can be seen as a subset of the positive operators $PO(n)$. $PO(n)$ has a natural partial order in the form of the Löwner order, which is trivial on $DO(n)$.
We can again consider a ``feature map'' $F:DO(n)\rightarrow PO(n)$ which possibly gives rise to a partial order $\rho \sleq \pi \iff F(\rho)\sleq_L F(\pi)$ where $\sleq_L$ is the Löwner order. For instance, setting $F(\rho)=\rho/\rho^+$ where $\rho^+$ is the highest eigenvalue of $\rho$ is the natural extension to the density operators of the first renormalised Löwner order described above. In fact, since the Löwner order restricted to diagonal matrices is equal to the product order on $\mathbb{R}^k$, this is a natural generalisation of the construction of information orders on $\Delta^n$.

\section{Conclusion and Further Research}
We have shown that there is a wide variety of partial orders on the space of probability distributions that satisfy the necessary conditions to capture the notion of information content. With an extra restriction (the degeneracy condition) we can make sure that this notion is unique. Unfortunately in practical linguistic applications this condition might prove to be too strict. The renormalised Löwner orderings are less strict in what they can compare and might prove to be more useful, although empirical research is needed to confirm this. The construction of the restricted information orders also points towards a way to create information orderings on the space of density operators, but studying this in detail is outside of the scope of this paper.

In the pursuit of methods that make comparisons between distributions easier we might look at rescaling distributions to study graded entailment (a generalisation of the approach taken in \cite{bankova2016}).
Another avenue of attack that might work is using the fact that in a high dimensional space words are probably far apart, so that we can be less picky with the comparisons, and set $x\sleq y$ whenever some elements within a certain radius of $x$ and $y$ are comparable. This procedure would break antisymmetry when considering the entire space, but not when only comparing words (assuming they are far enough apart). Doing this might allow elements in different sectors to be compared by a restricted information order.

\section*{Acknowledgements}
The author would like to thank the anonymous reviewers for their valuable feedback. Thanks goes out as well to the authors Master thesis supervisor Bob Coecke of which this publication is part, and to Martha Lewis and Daniel Marsden for valuable comments.

\bibliographystyle{eptcs}
\bibliography{cite}

\begin{thebibliography}{10}
\providecommand{\bibitemdeclare}[2]{}
\providecommand{\surnamestart}{}
\providecommand{\surnameend}{}
\providecommand{\urlprefix}{Available at }
\providecommand{\url}[1]{\texttt{#1}}
\providecommand{\href}[2]{\texttt{#2}}
\providecommand{\urlalt}[2]{\href{#1}{#2}}
\providecommand{\doi}[1]{doi:\urlalt{http://dx.doi.org/#1}{#1}}
\providecommand{\bibinfo}[2]{#2}

\bibitemdeclare{article}{balkir2015}
\bibitem{balkir2015}
\bibinfo{author}{Esma \surnamestart Balkir\surnameend},
  \bibinfo{author}{Dimitri \surnamestart Kartsaklis\surnameend} \&
  \bibinfo{author}{Mehrnoosh \surnamestart Sadrzadeh\surnameend}
  (\bibinfo{year}{2015}): \emph{\bibinfo{title}{Sentence Entailment in
  Compositional Distributional Semantics}}.
\newblock {\sl \bibinfo{journal}{arXiv preprint arXiv:1512.04419}}.
\newblock \urlprefix\url{https://arxiv.org/abs/1512.04419}.

\bibitemdeclare{incollection}{balkir2016}
\bibitem{balkir2016}
\bibinfo{author}{Esma \surnamestart Balkir\surnameend},
  \bibinfo{author}{Mehrnoosh \surnamestart Sadrzadeh\surnameend} \&
  \bibinfo{author}{Bob \surnamestart Coecke\surnameend} (\bibinfo{year}{2016}):
  \emph{\bibinfo{title}{Distributional sentence entailment using density
  matrices}}.
\newblock In: {\sl \bibinfo{booktitle}{Topics in Theoretical Computer
  Science}}, \bibinfo{publisher}{Springer}, pp. \bibinfo{pages}{1--22},
  \doi{10.1007/978-3-319-28678-5\_1}.

\bibitemdeclare{article}{bankova2016}
\bibitem{bankova2016}
\bibinfo{author}{Desislava \surnamestart Bankova\surnameend},
  \bibinfo{author}{Bob \surnamestart Coecke\surnameend},
  \bibinfo{author}{Martha \surnamestart Lewis\surnameend} \&
  \bibinfo{author}{Daniel \surnamestart Marsden\surnameend}
  (\bibinfo{year}{2016}): \emph{\bibinfo{title}{Graded Entailment for
  Compositional Distributional Semantics}}.
\newblock {\sl \bibinfo{journal}{arXiv preprint arXiv:1601.04908}}.
\newblock \urlprefix\url{https://arxiv.org/abs/1601.04908}.

\bibitemdeclare{inproceedings}{Baroni2012}
\bibitem{Baroni2012}
\bibinfo{author}{Marco \surnamestart Baroni\surnameend},
  \bibinfo{author}{Raffaella \surnamestart Bernardi\surnameend},
  \bibinfo{author}{Ngoc-Quynh \surnamestart Do\surnameend} \&
  \bibinfo{author}{Chung-chieh \surnamestart Shan\surnameend}
  (\bibinfo{year}{2012}): \emph{\bibinfo{title}{Entailment Above the Word Level
  in Distributional Semantics}}.
\newblock In: {\sl \bibinfo{booktitle}{Proceedings of the 13th Conference of
  the European Chapter of the ACL}}, \bibinfo{series}{EACL '12}, pp.
  \bibinfo{pages}{23--32}.
\newblock \urlprefix\url{http://dl.acm.org/citation.cfm?id=2380816.2380822}.

\bibitemdeclare{inproceedings}{clark2008}
\bibitem{clark2008}
\bibinfo{author}{Stephen \surnamestart Clark\surnameend}, \bibinfo{author}{Bob
  \surnamestart Coecke\surnameend} \& \bibinfo{author}{Mehrnoosh \surnamestart
  Sadrzadeh\surnameend} (\bibinfo{year}{2008}): \emph{\bibinfo{title}{A
  compositional distributional model of meaning}}.
\newblock In: {\sl \bibinfo{booktitle}{Proceedings of the Second Quantum
  Interaction Symposium (QI-2008)}}, pp. \bibinfo{pages}{133--140}.
\newblock
  \urlprefix\url{http://citeseerx.ist.psu.edu/viewdoc/download?doi=10.1.1.163.5435&rep=rep1&type=pdf}.

\bibitemdeclare{article}{coecke2010}
\bibitem{coecke2010}
\bibinfo{author}{B.~\surnamestart Coecke\surnameend},
  \bibinfo{author}{M.~\surnamestart Sadrzadeh\surnameend} \&
  \bibinfo{author}{S.~\surnamestart Clark\surnameend} (\bibinfo{year}{2010}):
  \emph{\bibinfo{title}{{M}athematical {F}oundations for a {C}ompositional
  {D}istributional {M}odel of {M}eaning. {L}ambek {F}estschrift}}.
\newblock {\sl \bibinfo{journal}{Linguistic Analysis}} \bibinfo{volume}{36},
  pp. \bibinfo{pages}{345--384}.

\bibitemdeclare{article}{coecke2003}
\bibitem{coecke2003}
\bibinfo{author}{Bob \surnamestart Coecke\surnameend} (\bibinfo{year}{2003}):
  \emph{\bibinfo{title}{Entropic geometry from logic}}.
\newblock {\sl \bibinfo{journal}{Electronic Notes in Theoretical Computer
  Science}} \bibinfo{volume}{83}, pp. \bibinfo{pages}{39--53},
  \doi{10.1016/S1571-0661(03)50003-2}.

\bibitemdeclare{inbook}{coecke2010book}
\bibitem{coecke2010book}
\bibinfo{author}{Bob \surnamestart Coecke\surnameend} \& \bibinfo{author}{Keye
  \surnamestart Martin\surnameend} (\bibinfo{year}{2011}):
  \emph{\bibinfo{title}{New Structures for Physics}}, chapter
  \bibinfo{chapter}{A Partial Order on Classical and Quantum States}, pp.
  \bibinfo{pages}{593--683}.
\newblock \bibinfo{publisher}{Springer Berlin Heidelberg},
  \bibinfo{address}{Berlin, Heidelberg}, \doi{10.1007/978-3-642-12821-9\_10}.

\bibitemdeclare{inproceedings}{kartsaklis2012}
\bibitem{kartsaklis2012}
\bibinfo{author}{Dimitri \surnamestart Kartsaklis\surnameend},
  \bibinfo{author}{Mehrnoosh \surnamestart Sadrzadeh\surnameend} \&
  \bibinfo{author}{Stephen \surnamestart Pulman\surnameend}
  (\bibinfo{year}{2012}): \emph{\bibinfo{title}{A Unified Sentence Space for
  Categorical Distributional-Compositional Semantics: Theory and Experiments}}.
\newblock In: {\sl \bibinfo{booktitle}{{COLING} 2012, 24th International
  Conference on Computational Linguistics, Proceedings of the Conference:
  Posters, 8-15 December 2012, Mumbai, India}}, pp. \bibinfo{pages}{549--558}.
\newblock \urlprefix\url{http://aclweb.org/anthology/C/C12/C12-2054.pdf}.

\bibitemdeclare{article}{kotlerman2010}
\bibitem{kotlerman2010}
\bibinfo{author}{Lili \surnamestart Kotlerman\surnameend}, \bibinfo{author}{Ido
  \surnamestart Dagan\surnameend}, \bibinfo{author}{Idan \surnamestart
  Szpektor\surnameend} \& \bibinfo{author}{Maayan \surnamestart
  Zhitomirsky-Geffet\surnameend} (\bibinfo{year}{2010}):
  \emph{\bibinfo{title}{Directional distributional similarity for lexical
  inference}}.
\newblock {\sl \bibinfo{journal}{Natural Language Engineering}}
  \bibinfo{volume}{16}(\bibinfo{number}{04}), pp. \bibinfo{pages}{359--389},
  \doi{10.1017/S1351324910000124}.

\bibitemdeclare{article}{martinphd}
\bibitem{martinphd}
\bibinfo{author}{Keye \surnamestart Martin\surnameend} (\bibinfo{year}{2000}):
  \emph{\bibinfo{title}{A foundation for computation}}.
\newblock {\sl \bibinfo{journal}{Tulane University, New Orleans, LA}}.
\newblock \urlprefix\url{http://www.nearmidnight.com/thesis.pdf}.

\bibitemdeclare{inproceedings}{piedeleu2015}
\bibitem{piedeleu2015}
\bibinfo{author}{Robin \surnamestart Piedeleu\surnameend},
  \bibinfo{author}{Dimitri \surnamestart Kartsaklis\surnameend},
  \bibinfo{author}{Bob \surnamestart Coecke\surnameend} \&
  \bibinfo{author}{Mehrnoosh \surnamestart Sadrzadeh\surnameend}
  (\bibinfo{year}{2015}): \emph{\bibinfo{title}{Open {S}ystem {C}ategorical
  {Q}uantum {S}emantics in {N}atural {L}anguage {P}rocessing}}.
\newblock In: {\sl \bibinfo{booktitle}{Proceedings of the 6th {C}onference on
  {A}lgebra and {C}oalgebra in {C}omputer {S}cience (CALCO)}},
  \bibinfo{address}{Nijmegen, Netherlands},
  \doi{10.4230/LIPIcs.CALCO.2015.270}.

\end{thebibliography}
\nocite{*}

\end{document}